\documentclass[letterpaper, 10 pt, conference]{ieeeconf}  %

\IEEEoverridecommandlockouts                              %

\usepackage{graphicx} %
\usepackage{amsmath} %
\usepackage{xcolor}
\usepackage{gensymb}
\usepackage[space]{cite}
\usepackage{siunitx}
\usepackage{hyperref}
\usepackage{relsize}

\DeclareSIUnit\px{px}
\DeclareSIUnit\min{min}

\title{\LARGE \bf
Leveraging Stereo-Camera Data for Real-Time Dynamic Obstacle Detection and Tracking
}

\author{Thomas Eppenberger$^{\star \dagger}$, Gianluca Cesari$^{\star}$, Marcin Dymczyk$^{\star}$, Roland Siegwart$^{\dagger}$, and Renaud Dub\'{e}$^{\star}$%
\thanks{$^{\star}$Sevensense Robotics AG, Zurich, Switzerland\newline \{firstname.lastname@sevensense.ch\}
}%
\thanks{$^{\dagger}$ETH Zurich, Autonomous Systems Lab (ASL), Zurich, Switzerland}
}

\begin{document}

\maketitle
\thispagestyle{empty}
\pagestyle{empty}

\begin{abstract}
Dynamic obstacle avoidance is one crucial component for compliant navigation in crowded environments. In this paper we present a system for accurate and reliable detection and tracking of dynamic objects using noisy point cloud data generated by stereo cameras. Our solution is real-time capable and specifically designed for the deployment on computationally-constrained unmanned ground vehicles.
The proposed approach identifies individual objects in the robot's surroundings and classifies them as either static or dynamic. The dynamic objects are labeled as either a person or a generic dynamic object. We then estimate their velocities to generate a 2D occupancy grid that is suitable for performing obstacle avoidance.
We evaluate the system in indoor and outdoor scenarios and achieve real-time performance on a consumer-grade computer.
On our test-dataset, we reach a MOTP of $\mathbf{0.07 \pm 0.07m}$, and a MOTA of $\mathbf{85.3\%}$ for the detection and tracking of dynamic objects. We reach a precision of $\mathbf{96.9\%}$ for the detection of static objects.

\end{abstract}

\vspace{-1mm}
\section{INTRODUCTION} \label{introduction}

In order to safely and efficiently navigate in crowded public spaces, Unmanned Ground Vehicles (UGVs) need to reason about their static and dynamic surroundings and predict the occupancy of space to reliably perform obstacle avoidance\cite{ess2009moving}.
Generally, static objects can be avoided by small safety distances, whereas
for compliant navigation, dynamic objects need to be avoided by larger distances~\cite{pfeiffer2016predicting}. Moreover, robots should avoid crossing pedestrian paths, which not only requires to correctly classify dynamic objects as such, but also calls for accurate motion estimation and prediction. In addition to humans, other dynamic objects with varying size and speed, such as animals or other vehicles, may appear in the surroundings. Hence, the detection may not be restricted to humans only but needs to generalize.

We identified two major families of state-of-the-art techniques for detecting and tracking objects in crowded scenes. The first group uses point cloud data, often generated from highly-accurate LiDAR sensors, which allows for the detection of generic dynamic objects and is typically used for autonomous driving~\cite{kraemer2018lidar, asvadi20163d}. The costs of LiDAR sensors make most of the algorithms in this first group not applicable for small, commercially used UGVs, such as delivery robots for hospitals or airports. The second group uses visual information from images and primarily focuses on the detection of a predetermined number of object classes, such as pedestrians and vehicles\cite{jafari2014real, redmon2018yolov3}. However, these approaches often lack the ability to detect generic dynamic objects and
do not run in real-time on computationally-constrained platforms~\cite{zhang2018towards}. In order to successfully deploy UGVs at large scale, low-cost sensor setups, such as stereo cameras, should be used along with efficient algorithms.

\begin{figure}[t]
  \centering
  \includegraphics[width=1.0\columnwidth]{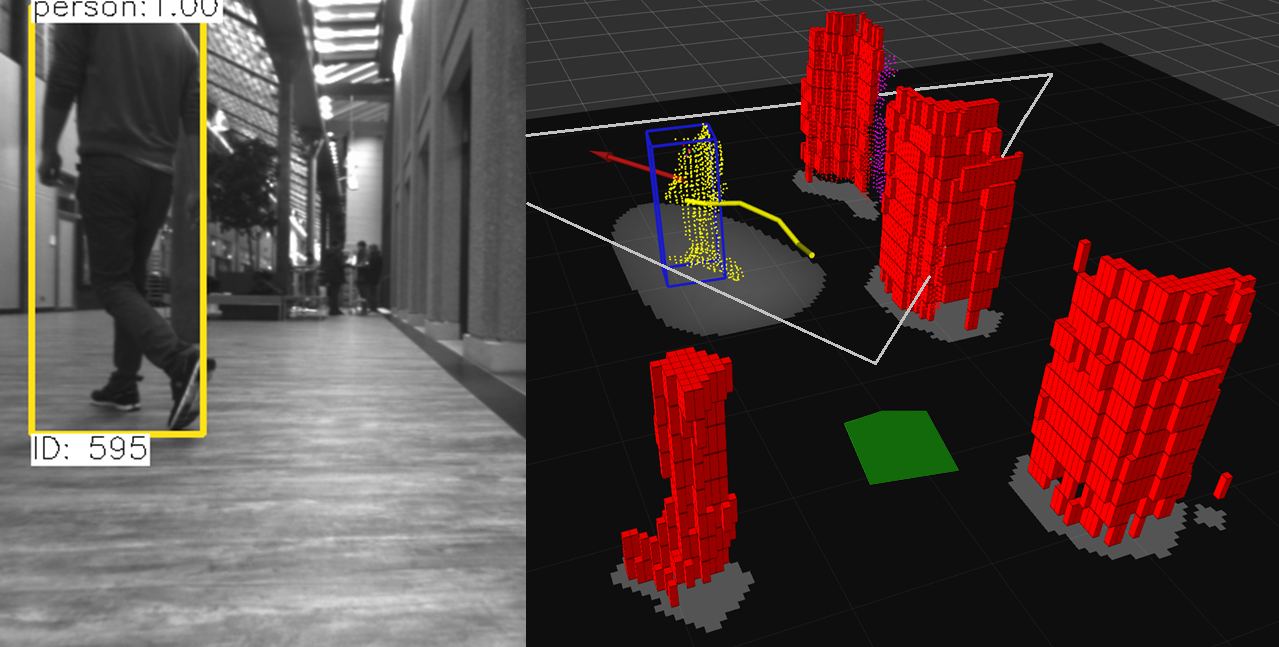}
  \vspace{-6mm}
  \caption{Visualization of the output of the proposed dynamic object detection and tracking approach.
  \textbf{Left:} the input camera image overlaid with the output of a visual people detector, indicating the confidence of the detection and a tracking ID. \textbf{Right:} a resulting occupancy grid, with correctly identified static objects (red voxels) and a detected pedestrian (the yellow point cloud). The red arrow visualizes the pedestrian's estimated velocity, the yellow track shows the past trajectory, and the blue cuboid indicates that the visual people detector recognized this cluster as a person. Gray floor areas indicate high costs in the occupancy grid.
  The robot's footprint and field of view are shown in green and white, respectively.}
  \label{fig_title_page}
  \vspace{-5mm}
\end{figure}

We introduce a
solution that leverages stereo camera data to reliably and accurately detect and track dynamic objects. To this end we first present a novel algorithm to detect generic dynamic objects based on their motion. For enhanced perceptual performance in crowded spaces, we use a visual people detector to classify humans motion-independently as a specific class of dynamic objects, as depicted in Figure~\ref{fig_title_page}. Our approach handles short-time occlusions using the estimated velocity of the dynamic objects.
To the best of our knowledge this is the first work to propose a complete solution that uses stereo cameras for detecting and tracking generic dynamic objects by combining global nearest neighbor searches and a visual people detector.
The system relies on noisy data of one stereo camera only and is designed to run on computationally-constrained platforms. As shown in the work of Liu \cite{LuciaQiLiu2020navigation}, our perception system has been used for navigating an UGV in real life crowds.
We encourage the reader to consult the supplementary video (\url{https://youtu.be/AYjgeaQR8uQ}) for more visualizations.

The contributions of this paper are as follows:

\begin{figure*}[t]
  \centering
  \includegraphics[width=\textwidth]{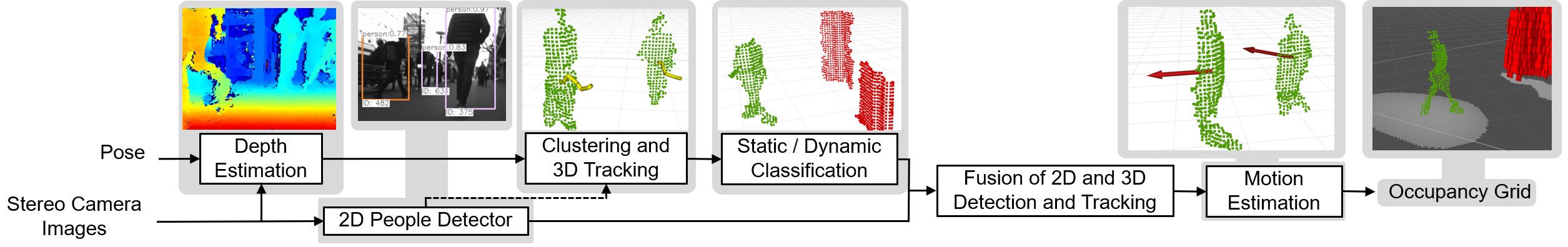}
  \vspace{-5mm}
  \caption{
  Overview of our pipeline:
  the inputs are stereo images and the estimated pose of the robot from a visual SLAM module.
  The output is a 2D occupancy grid, which enables planning paths close to static objects and ensures avoidance of dynamic objects by a safety distance.}
  \label{fig_pipeline}
  \vspace{-4.3mm}
\end{figure*}

\begin{itemize}
    \item a novel real-time algorithm to detect and track generic dynamic objects based on noisy stereo camera-data;
    \item a method to combine the aforementioned algorithm with a vision-based people detector to improve the detection and tracking performance and handle \mbox{short-time occlusions};
    \item an evaluation of our pipeline on challenging datasets, demonstrating its performance and reliability for increased mobile robot safety.
\end{itemize}

\section{RELATED WORK} \label{related_work}

The task of detecting and tracking dynamic agents has led to a variety of different approaches, as robotic platforms range from small, commercial UGVs to autonomous cars. These approaches differ in computational load, robustness against noise, and required sensor data, namely point clouds, images, or a combination of both.

\textbf{Image based algorithms:}
Assuming that a feature-based SLAM system is used, one approach is to utilize outliers of the feature tracking module to detect dynamic objects~\cite{xu2018mid, barsan2018robust}. Song et al. \cite{song2017real} argue that this approach should be favored over the usage of optical flow to estimate the velocities of visible objects \cite{pfeiffer2010efficient}. This technique, however, requires the tracking algorithm to use dense feature points to ensure outliers point to all dynamic objects. This is not guaranteed for many visual SLAM systems and contradicts with our goal to design a generic dynamic object tracking pipeline.

Nonetheless, the optical flow methods can be enhanced to produce so-called Scene Flow, which computes per-pixel 3D velocities. However, this method does not run in real-time on computationally-constrained platforms~\cite{schuster2018combining, DBLP:journals/corr/abs-1808-09297}.

Another approach is the specific detection of pedestrians using visual data \cite{nguyen2019confidence, zhang2018towards, 
dollar2014fast, ess2009moving
}, where deep neural networks receive much attention \cite{simon2019complexer, redmon2018yolov3, zhang2016faster
}.
Segmentation networks \cite{siam2018rtseg, paszke2016enet} are an alternative to detector networks, but do not 
imply
object-instances, and hence, are impractical for differentiation between multiple people in crowded scenes. Object-instance segmentation networks overcome this drawback \cite{he2017mask}. In this work, we do not limit ourselves to detecting certain object classes only, but pursue a generic dynamic object detection.

\textbf{Point cloud based algorithms:}
detection algorithms based on Iterative Closest Point (ICP) match current segments of a point cloud to a previous point cloud to reveal their motion \cite{jiang2017high, li2017rgb, christie20163d}. This approach works best for rigid objects like cars and considerably decreases in performance for deforming objects like humans. This property does not match the needs of our system, where we put a special emphasis on human detection and tracking.

When using a volumetric occupancy grid,
the classification of points as static or dynamic can be based on the consistency of the occupancy of the voxels
\cite{asvadi20163d, azim2014layer, broggi2013full}. The performance of this approach depends on the voxel-size where smaller voxels allow for a more precise occupancy analysis, while requiring more computational effort. A probabilistic metric for the occupancy of the voxels is necessary to handle noise in the data, introducing a time-lag to the classification process.

Moreover, there are several algorithms designed for autonomous driving that rely on high-quality point clouds from LiDAR sensors and powerful GPUs to detect cars and pedestrians \cite{kraemer2018lidar, ku2018joint, miller2016dynamic}.
In contrast, our target platforms are small UGVs with low computational power and potentially the lack of an expensive LiDAR sensor.

Our approach directly compares point clouds among frames and is related to the work of Yoon et al.~\cite{yoon2019mapless} and Shi et al.~\cite{shi2018dynamic}. Yoon et al.~\cite{yoon2019mapless} work with LiDAR data and need to introduce a time-lag of one frame to reliably cope with occlusions. We expand their idea to handle noisy stereo camera data with potentially incomplete depth information and additionally introduce a novel approach to differentiate between dynamic and previously occluded points without the need of a time-lag of one frame. Shi et al.~\cite{shi2018dynamic} use RGB-D data and remove points during dense reconstruction in case they are spatially inconsistent between frames. In contrast, our method is able to classify all points of an object as dynamic, even though only their subset shows spatial inconsistency. In consequence, we obtain a more complete and robust classification of dynamic objects. Compared to the work of Osep et al. \cite{osep2017combined} we do not limit our system to a set of predefined detectable classes but implement a generic dynamic object detector instead.

\section{METHODOLOGY} \label{methodology}

An overview of the proposed stereo camera-based perception approach is given in Figure~\ref{fig_pipeline} and
the remainder of this section details its individual modules.
To associate camera-based point clouds with the global frame we localize the robot using precise Visual Inertial Odometry (VIO).

\subsection{Point Cloud Generation} \label{cloud_generation}

The first module generates a 3D point cloud from undistorted and rectified stereo images.
We designed our approach to be generic regarding the inputs, hence any algorithm extracting a disparity map from stereo images can be used in this module, from which we consider the well-established block-matching and deep neural networks.

\subsubsection{Block-Matching} \label{block_matching}

We use semi-global block-matching~\cite{hirschmuller2007stereo} and apply a weighted-least-squares filter \cite{min2014fast} on the resulting disparity map.

\subsubsection{Deep Stereo} \label{deep_stereo}

\begin{figure}[t]
  \centering
  \includegraphics[width=1.0\columnwidth]{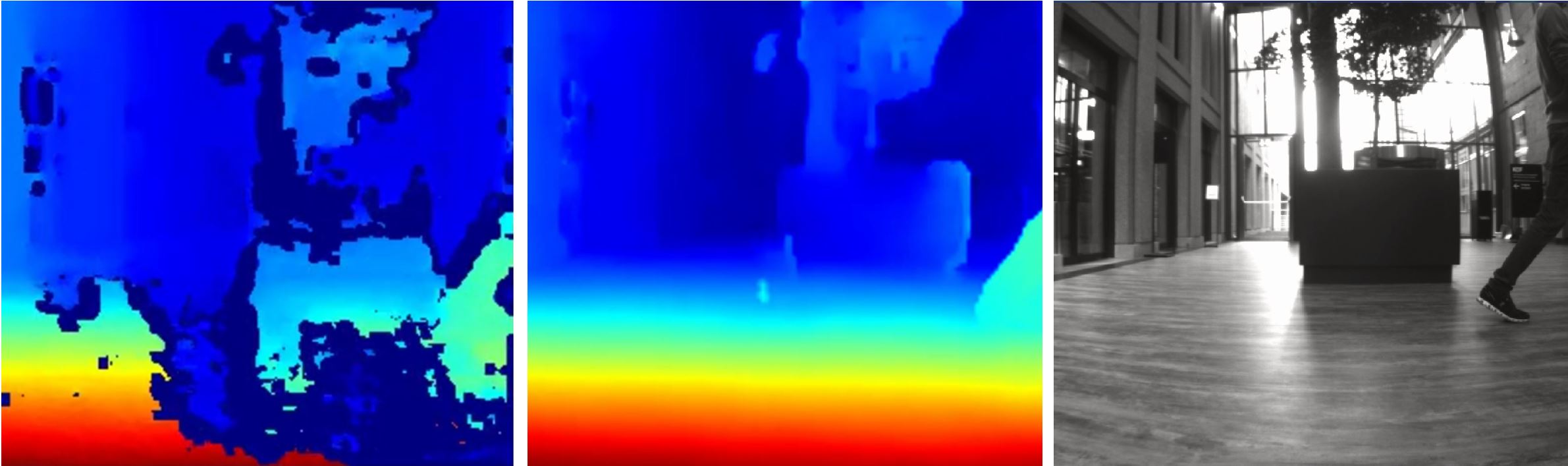}
  \caption{Depth representations generated using stereo images. \textbf{Left:} Block-matching~\cite{hirschmuller2007stereo} cannot generate depth information in parts of the low-textured object on the right, or on the shiny surface of the floor.
  \textbf{Middle:} MADNet \cite{tonioni2018real} captures most parts of the object and the floor. Hence, it delivers more complete depth information than block-matching in this scenario. \textbf{Right:} raw image.}
  \label{fig_deep}
  \vspace{-6mm}
\end{figure}

Recently, deep neural networks that learn to infer disparity values from stereo images have emerged~\cite{khamis2018stereonet, chang2018pyramid,
mayer2016large,
wang2018anytime}. We use MADNet \cite{tonioni2018real}, as we found this network to deliver a suitable trade-off between run-time and performance. Figure~\ref{fig_deep} shows an exemplary disparity map generated by both methods.

\subsection{Point Cloud Filtering} \label{filter_point_cloud}

We filter the point cloud generated by the previous module to reduce noise and down-sample the data for achieving real-time performance. We denote the point cloud after initial cropping as $h^d$ and after filtering as $h^s$. We applying the following sequence:

\begin{itemize}
    \item crop the point cloud at the depth limit $l_{d}$, up to which the measurements can be trusted\footnote{The limit $l_{d}$ must be chosen based on the camera setup and the algorithm generating the disparity map.};
    \item crop the point cloud at the heights $l_{g}$ and $l_h$, to remove all ground plane and ceiling points, respectively.
    \item apply a voxel filter with leaf size $l_{l}$
    to reduce the size of the point cloud by an order of magnitude and to ensure even density of the points in 3D.
    \item apply a
    filter to remove all points with less than $l_n$ neighbors within a radius $l_r$, to reduce noise points which occur most notably at the edges of objects.
\end{itemize}

\subsection{Clustering and 3D Tracking} \label{clustering_tracking}

In this module we identify the individual objects through clustering and track them from frame to frame.

\subsubsection{Clustering} \label{clustering}

We use DBSCAN~\cite{ester1996density} to cluster the point cloud, resulting in a set of $m$ clusters $C=\{C^1, C^2, ..., C^m\}$. DBSCAN grows clusters from random seeds using dense points only.
Compared to Euclidean clustering \cite{douillard2011segmentation}, we experienced that DBSCAN more precisely separates individual objects in cluttered point clouds, while introducing only a marginal computational overhead.

To refine the clusters, we use the bounding-boxes generated by the 2D people detector module.
We separate any clusters which are associated with more than one bounding-box to distinguish individual humans.
We also separate clusters whose fraction of points laying within the bounding-box is below a threshold to distinguish between humans and nearby static objects.
In Section~\ref{people_detector} we describe how the bounding-boxes are obtained and associated to the clusters.

\subsubsection{3D tracking} \label{tracking}

First, at time $t$ we compute the centroids $c_t$ of all current clusters $C_t$ in the global frame as the average of all their points. Then, we associate them to their closest centroid $c_{t-\Delta t}^*$ of the clusters $C_{t-\Delta t}$ of the previous frame. Applying the tracking over $k$ frames separated by $\Delta t$ leads to a cluster track $T_{t,k}^i=\{c_{t-k\cdot \Delta t}^*, ..., c_{t-\Delta t}^*, c_{t}^i\}$ for a current cluster $C_t^i$. We mark current centroids $c_t$ that cannot be related to any previous centroids $c_{t-\Delta t}^*$ as newly appeared objects and non-connected previous centroids as lost objects.

\begin{figure}[t]
  \centering
  \includegraphics[width=1.0\columnwidth]{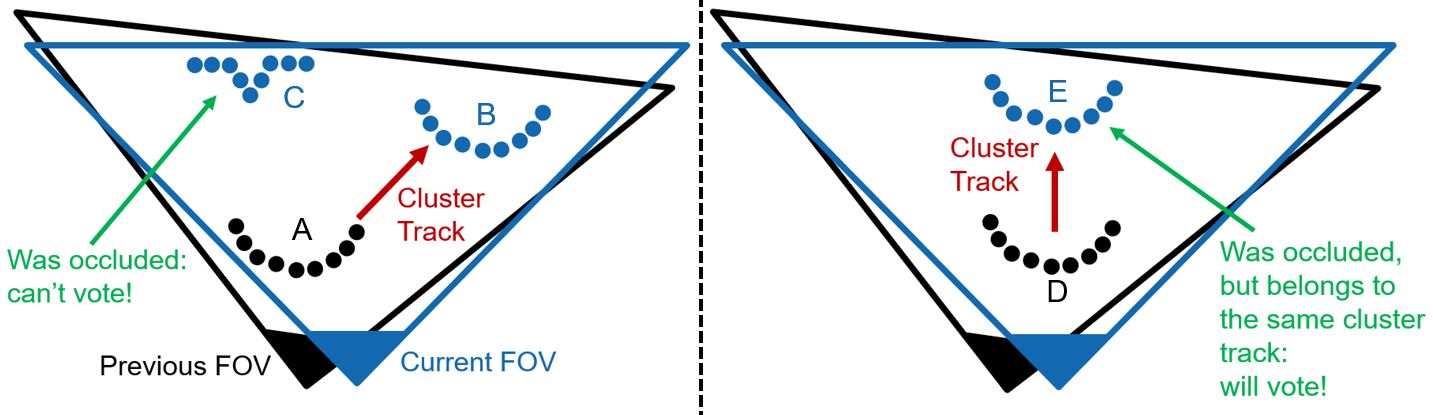}
  \caption{Occlusion handling at dynamic object detection. \textbf{Left:} the current cluster \textit{C} is excluded from voting, as it was occluded in the previous frame by cluster \textit{A}, which belongs to a different cluster track $T_{t,1}^{B}$. \textbf{Right:} the current cluster \textit{E} is \textit{not} excluded from voting, as in the previous frame it was occluded by cluster \textit{D}, which belongs to the same cluster track $T_{t,1}^{E}$.}
  \label{fig_occlusion_explained}
  \vspace{-6mm}
\end{figure}

\subsection{Classification as Dynamic or Static} \label{classification}

In this module the clusters $C_t^i$ identified in Section~\ref{clustering_tracking} are classified as either static or dynamic based on a voting strategy. First, we let the individual points of a cluster vote for the cluster's class.
Then, the classification of the cluster is based on the voting information of all its points. Namely, we classify it as dynamic if the absolute or relative amount of votes for being dynamic surpass the respective thresholds $l_{dyn}^{abs}$ or $l_{dyn}^{rel}$. We use two thresholds in order to correctly classify objects at different scales.
In case the classification is inconsistent over a short time horizon $\tau$, we mark the cluster as uncertain.
Every cluster classified as dynamic is regarded as being an individual object.

Below, we describe the voting process of an individual point and indicate which points are excluded from voting.

\subsubsection{Voting of an individual point} \label{voting}

We measure the global nearest neighbor distance $d^k$ 
from each point $k$ of a cluster $C_t^i$ in the current \textit{filtered} point cloud $h^s$ to a previous \textit{dense, non-filtered} point cloud $h^d_{t-\delta}$. We found that it is key to measure $d^k$ from the \textit{filtered} to the \textit{dense, non-filtered} point cloud $h^d_{t-\delta}$ in order to gain robustness against noise.

In order to assure that points of dynamic objects move substantially more than points corrupted by noise, we compare point clouds from frames being roughly $\delta$ seconds apart, similarly to the work of Yoon et al. \cite{yoon2019mapless}.
For
static objects, these measured nearest neighbor distances $d$ will be in the magnitude of the noise. For points at the leading edge of moving objects, however, $d$ will be substantially higher.
We convert these distances to velocities and let each point $k$ with $\frac{d^k}{\delta} < l_{NN}$ vote as being static, while points with $\frac{d^k}{\delta} \geq l_{NN}$ will vote as being dynamic. $l_{NN}$ denotes the velocity threshold, which needs to be set marginally higher than the noise level in the filtered point cloud $h^s$.

\subsubsection{Excluding points from voting} \label{exclude_voting}

We only can infer knowledge about a point's movement if we could observe it in both frames used for voting. This observability-requirement is not fulfilled in two cases which we detect for improving the voting performance.

First, if the Field of View (FOV) of a robot changes between the two frames, points might appear in the area of the current FOV that does not overlap with the previous FOV. As we have observed those points only once, we exclude them from voting.

Second, we exclude points of a current cluster $C_t^i$ from voting if they previously were occluded by a \textit{different} object $j$, i.e. by points from a cluster~$C^j_{t-\delta}$ from a different cluster track $T_{t,k}^{j\neq i}$. Specifically, we distinguish between such occlusions and self-occlusions which happen when objects move away from the camera, as visualized in Figure \ref{fig_occlusion_explained}.
Occlusions are identified by first approximating the depth map of the previous frame $g_{t-\delta}$ by projecting randomly-sampled points of the non-filtered, dense point cloud $h^d_{t-\delta}$ onto the previous image plane.
We then project a query point $q_t \in C_t^i$ onto $g_{t-\delta}$ and run a 2D nearest neighbor search. If a close nearest neighbor ${n_{t-\delta}^{2D} \in g_{t-\delta}}$ is found, we check for a potential occlusion, that is $depth[n_{t-\delta}] < depth[q_t]$, with $n_{t-\delta}$ being the associated 3D point of $n_{t-\delta}^{2D}$.
To identify self-occlusions we associate $n_{t-\delta}$ to the cluster track of its nearest neighbor in $h_{t-\delta}^s$ and check if $T_{t-\delta}^{n_{t-\delta}} \in T_{t}^{q_t}$. In this case, the query point $q_t$ is \textit{not} excluded from voting.

\subsection{2D People Detector} \label{people_detector}

\begin{figure}[t]
  \centering
  \includegraphics[width=0.8\columnwidth]{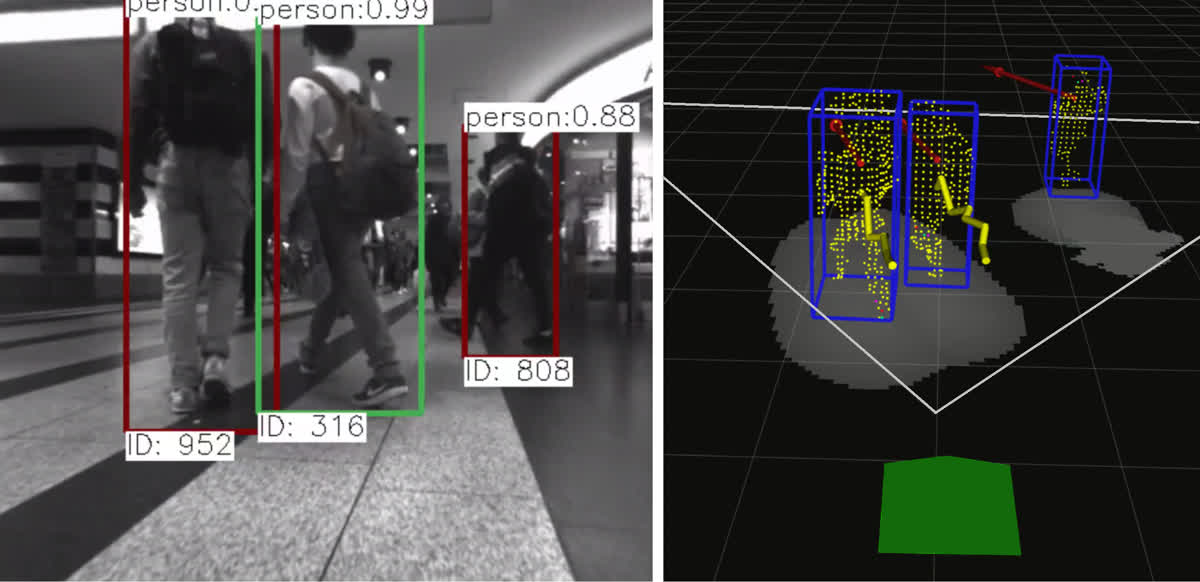}
  \caption{\hspace{-1mm}\textbf{Left}: a sample output of the Mobilenet-SSD people detector\cite{howard2017mobilenets, liu2016ssd}. Every detection is rated by a confidence, shown on top. We assign an ID to every bounding-box being tracked on the image plane. \textbf{Right}: we associate detections to clusters, indicated by the blue cuboids.}
  \label{fig_detector}
  \vspace{-5mm}
\end{figure}

Up to now,
the system would classify a standing person as static and would only realize that it was a dynamic object once she starts walking. By adding Mobilenet-SSD~\cite{howard2017mobilenets, liu2016ssd} as a visual 2D people detector to our pipeline, we achieve a motion-independent detection of pedestrians. We select this network as it
delivers a suitable trade-off between run-time and performance on grayscale images.

Figure~\ref{fig_detector} shows an exemplary output of the Mobilenet-SSD people detector. %
We track the bounding-boxes over frames using Intersection over Union (IoU) as a metric. We use tracking by detection, as visual trackers have not yet reached a satisfying performance level while running in real-time on
CPUs~\cite{leal2017tracking}. Similar to the cluster tracks in 3D, we generate bounding-box tracks $B_{t,k}$ in the image plane. In the subsequent Section~\ref{fusion}, we use $B_{t,k}$ to make the 3D tracking more robust.

The 2D bounding-boxes are associated to the 3D clusters $C$ by linking the detection to the cluster having the highest amount of points within the bounding-box.
Note that other approaches exist for this 2D-3D association~\cite{shi2018dynamic, ku2018joint, qi2018frustum, lahoud20172d}.

\subsection{Fusion of 2D and 3D Detection and Tracking} \label{fusion}

The inputs to this module are the 3D cluster tracks $T_{t}$ classified as static or dynamic, and the 2D bounding-box tracks $B_{t}$.

For every cluster track $T_{t}^i$, the frequency $f_a^i$ of associated 2D-to-3D detections is computed. If $f_a^i$ is above a certain confidence-threshold $\gamma$, we classify all clusters being added to this cluster-track as representing a pedestrian, and hence, representing a dynamic object regardless of their initial classification.
Furthermore, this module checks if all bounding-boxes of a track $B_{t}^j$ are consistently associated to clusters of the same cluster track $T_{t}^i$, and resets $f_a^i$ otherwise.

\subsection{Motion Estimation} \label{motion_estimation}

\def\A{
\begin{bmatrix}
    1 & 0 & T_s &0 \\
    0 & 1 & 0 & T_s \\
    0 & 0 & 1 & 0 \\
    0 & 0 & 0 & 1 \\
\end{bmatrix}}

Estimating the velocity and predicting the future trajectory of pedestrians is an active research field \cite{pfeiffer2018data, chen2017socially, wulfmeier2017large, pfeiffer2016predicting}.
Similarly to the work of Azim et al.~\cite{azim2012detection}, we adopt a conservative motion model to estimate the velocities and short-term future paths of dynamic objects.
Assuming that the dynamic objects move on a horizontal plane, we estimate their velocity by using a constant 2D velocity model, based on a Kalman filter (KF).
The measurement inputs $\vec{z}_i$ for the KF are the cluster centroids $\vec{c}_i = [c_x, c_y, c_z]_i^\top$ of a cluster track $T_{t}^i$ measured in the x-y plane of the world frame: $\vec{z}_i = [c_x, c_y]_i^\top.$
We define the state vector as: $\vec{x}_i = [x, y, \dot{x}, \dot{y}]_i^\top$.
The system dynamics and measurement model are defined as:
$$\vec{x}_i[k+1] = A[k] \cdot \vec{x}_i[k] + N(0, Q)$$
$$\vec{z}_i[k] = H[k] \cdot \vec{x}_i[k] + N(0, R)$$
where $Q$ and $R$ model the system noise and measurement noise, respectively. $H$ extracts the first two dimensions of $\vec{x}$ and $A$ is defined as:
$$A = \A$$
where $T_s$
denotes the time between two updates. Using the KF, we can catch short-time occlusions of dynamic objects. Specifically, when an object $i$ is lost during tracking, we keep the KF running and compute for all new appearing objects $j$ the probability $p(j=i)$ of being the same object as the lost one. We compute this probability as ${p(j=i) = N(\vec{c}_j | \vec{c}_i, C_i(\vec{x}_i))}$, with $C_i(\vec{x}_i)$ being the estimated covariance of the state $\vec{x}_i$. We connect the cluster tracks of object $i$ and $j$ if $p(j=i)$ surpasses a threshold.

\subsection{2D Occupancy grid} \label{costmap}

In order to allow for path planning and obstacle avoidance, we leverage occupancy grid representations~\cite{oleynikova2016voxblox, fankhauser2016universal, hornung2013octomap} and specifically use the computationally efficient Costmap\_2d implementation \cite{lu2014layered}. 
We create three maps in parallel for static, dynamic and uncertain objects, in which the obstacles are represented as positive costs at their respective location.
In the uncertain map, we create only short-living costs for which no static/dynamic classification was done yet. In the static map,
we use raytracing \cite{marder2010office} to clear free space once it was erroneously occupied. In the dynamic map, we expand the costs in the direction of the estimated velocity of the objects, such that a robot will not collide with it in the future.
Safe paths can then be planned and executed by aggregating the costs of the three aforementioned layers. 

\section{EVALUATION} \label{evaluation}

In this section, we evaluate the performance of the presented obstacle classification and tracking solution. The run-times of the individual modules are finally presented.
\subsection{Experimental Setup} \label{experimental_setup}

\begin{figure}[t]
  \centering
  \includegraphics[width=0.8\columnwidth]{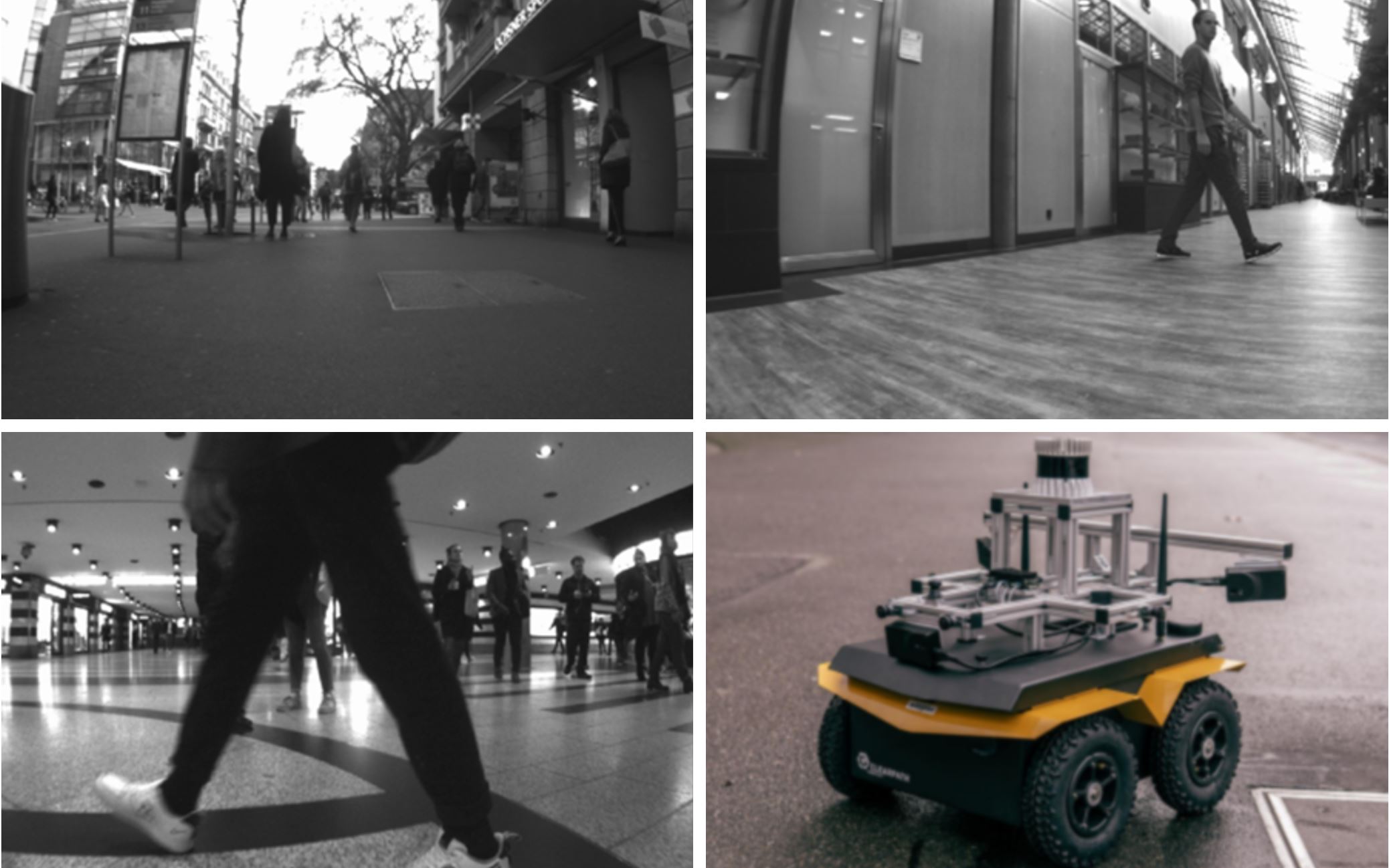}
  \caption{Sample images used in our evaluation and the platform we used to collect them. Our system relies on the stereo camera mounted in the front, whereas the LiDAR was used for evaluation purposes only.
  }
  \label{fig_datasets}
  \vspace{-5mm}
\end{figure}

We recorded multiple stereo-vision datasets that include pedestrians, featuring challenging indoor and outdoor scenes, shiny surfaces, low-textured objects, illumination changes, and empty as well as crowded spaces. To record our datasets, we used a Clearpath Robotics
Jackal robot equipped with a stereo camera
(grayscale, $752\times 480\si{\px}$),
and an Ouster OS-1 LiDAR with 16 channels for the ground truth measurements used in Section~\ref{exactness_completeness} and \ref{classification_accuracy}. Sample images of our datasets and our robot are shown in Figure~\ref{fig_datasets}. For all experiments, including the timing presented in Section~\ref{timing}, we run our pipeline on a
Intel i$7$-$8650$U processor.

For our stereo camera setup we set the parameters introduced in Section~\ref{filter_point_cloud} as ${l_{l} = 0.05m}$, ${l_{g} = 0.15m}$, ${l_{h} = 1.8m}$, ${l_{n} = 30}$, ${l_r = 0.5m}$ %
and $\delta = 0.4s$. Furthermore, we set ${l_{d} = 5m}$, as we chose a disparity-error of $1\si{\px}$ leading to a depth-error of $0.5m$ to be the limit we can accept. Regarding the parameters of Section~\ref{clustering_tracking}, \ref{classification}, and \ref{fusion} we set $l_{dyn}^{abs} = 100$, $l_{dyn}^{rel} = 0.8$, ${\tau = 0.4s}$, ${l_{NN} = 0.45m/s}$, and $\gamma = 1.5s^{-1}$. The robot's nominal speed is $1.2m/s$. We verified that our settings are
sufficient for the objects
to be reliably tracked
up to the speed of a jogging person.
Note that these parameters were identified using datasets independent of those used in the evaluations.

For precise localization of our robot we run a 
VIO 
algorithm similar to OKVIS \cite{leutenegger2015keyframe}. For image processing we use OpenCV\cite{opencv_library}. The people detector Mobilenet-SSD \cite{howard2017mobilenets, liu2016ssd} is implemented in Caffe and the deep stereo network MADNet \cite{tonioni2018real} in Tensorflow. 

\subsection{Accuracy and Completeness of the Point Clouds} \label{exactness_completeness}

\begin{figure}[t]
\vspace{-7mm}
  \centering
  \hbox{\hspace{1.9mm} \includegraphics[width=1.0\columnwidth]{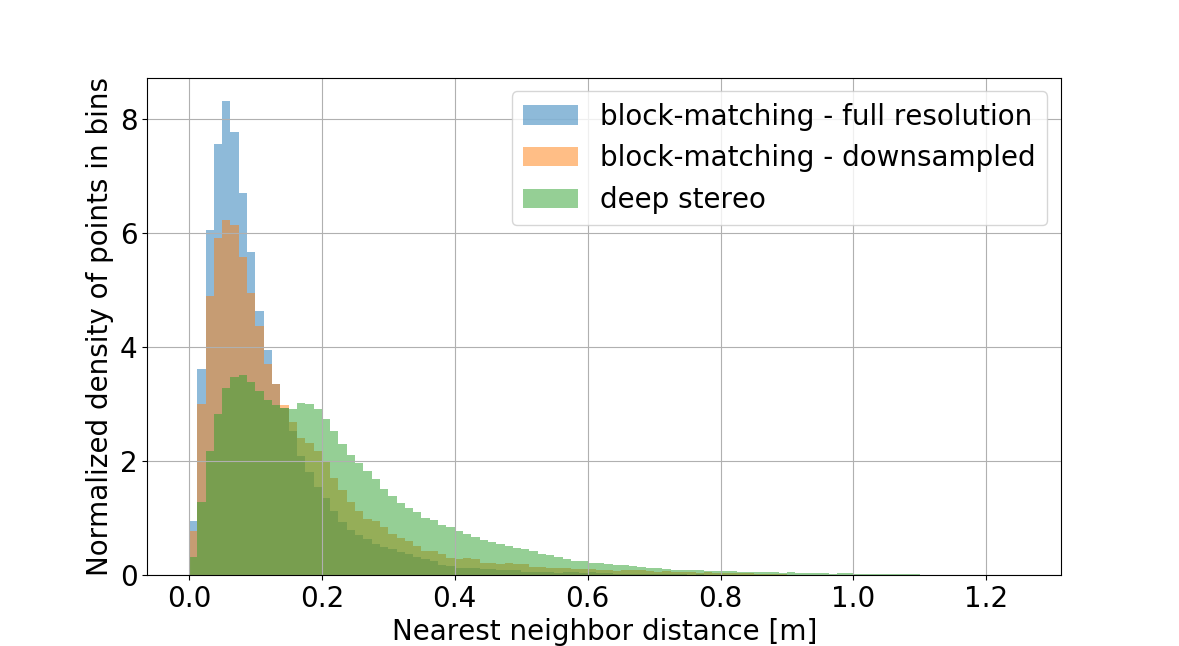}}
  \includegraphics[width=0.9\columnwidth]{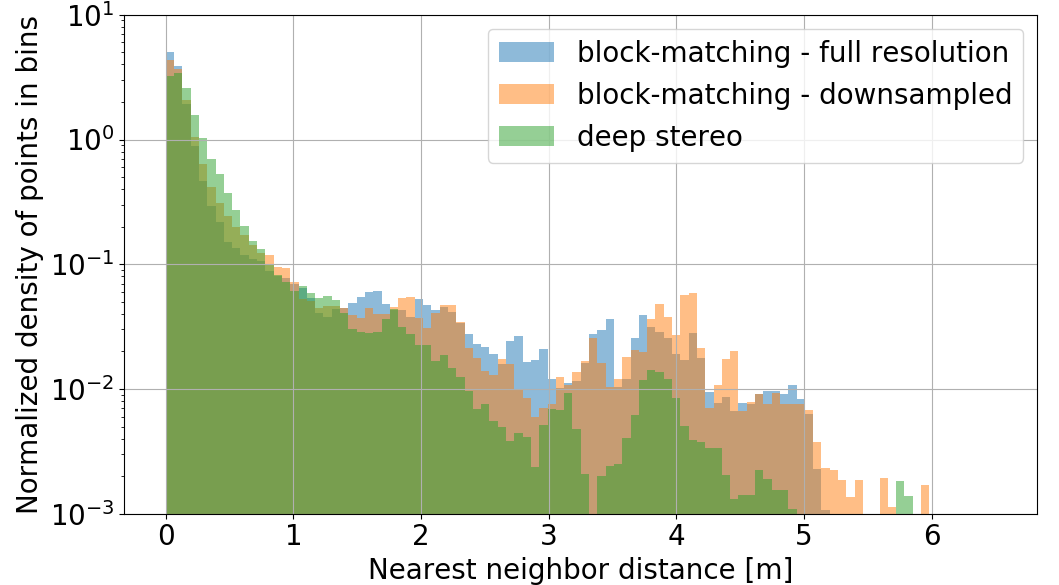}
  \caption{Normalized histograms of nearest neighbor distances~$d$ between point clouds from the LiDAR and the stereo camera to analyze the \textit{accuracy} and \textit{completeness}. \textbf{Top/Accuracy:} $d$ measured 
  from the camera to the LiDAR.
  \textbf{Bottom/Completeness:} $d$ measured
  from the LiDAR to the camera.
  The region above the accuracy-limit $l_c = 0.8m$ indicates points of objects not captured by the stereo camera point clouds.
  }
  \label{fig_exactness_completeness}
  \vspace{-3mm}
\end{figure}

In this section, we briefly evaluate the quality of the stereo camera point cloud to identify the accuracy-limit $l_c$, which we will use in Section \ref{classification_accuracy} to evaluate the static object detection precision.

In order to evaluate the performance of the two point cloud generation methods identified in Section~\ref{cloud_generation}, we collected  ground truth 3D LiDAR data synchronized with the stereo images.
To compare both the LiDAR and vision point clouds, we use nearest neighbor distances~$d$ as the metric, that is $d = || p_1 - p_2 ||_2$, for two
points $p_1$ and $p_2$. 
The \textit{accuracy} of our stereo camera point clouds is estimated by computing $d$ from each point of the camera clouds to its nearest neighbor in the LiDAR clouds.
In the opposite manner, we measure the \textit{completeness} of our camera clouds by computing $d$ from each point of the LiDAR clouds to its nearest neighbor in the camera clouds. 
Note that, as the LiDAR clouds do not feature any measurements between beams, there will be non-zero nearest neighbor distances $d$, even in an ideal setting. In our setup\footnote{A 16-channels LiDAR with an opening angle of $30\degree$ and $l_d = 5m$.}, this distance is at most $0.08m$.

We evaluate the accuracy and completeness of the camera point cloud on a dataset where we drove down a crowded public sidewalk. In order to achieve real-time capability, we simplify the block-matching process by down-sampling images with a factor of two.

\subsubsection{Accuracy}

The histogram depicted in Figure~\ref{fig_exactness_completeness} shows the normalized distribution of the
distances $d$.
The block-matching point cloud performs best and has an accuracy-limit of $l_c = 0.8m$, as all point are below this value. Clearly, the deep stereo network cannot compete with block-matching. Please refer to MADNet \cite{tonioni2018real} for in-depth analysis of the performance of this network.

\subsubsection{Completeness}

The histogram of the normalized distribution of the
distances $d$ is shown in Figure~\ref{fig_exactness_completeness}. Measurements substantially larger than $l_c = 0.8m$ indicate points of objects that were not captured by the camera cloud.
Deep stereo performs best, which was suggested by Figure~\ref{fig_deep}. Overall, we observe that deep stereo is less exact, but more complete than block-matching. More detailed comparisons of methods for generating dense maps from stereo data can be found in the work of Menze \cite{menze2015object}.
For the remaining parts of the evaluation, we use block-matching on downsampled images, due to its real-time capability and accuracy.

\subsection{Classification and Tracking Accuracy} \label{classification_accuracy}

\begin{figure}[t]
  \centering
  \includegraphics[width=1.0\columnwidth]{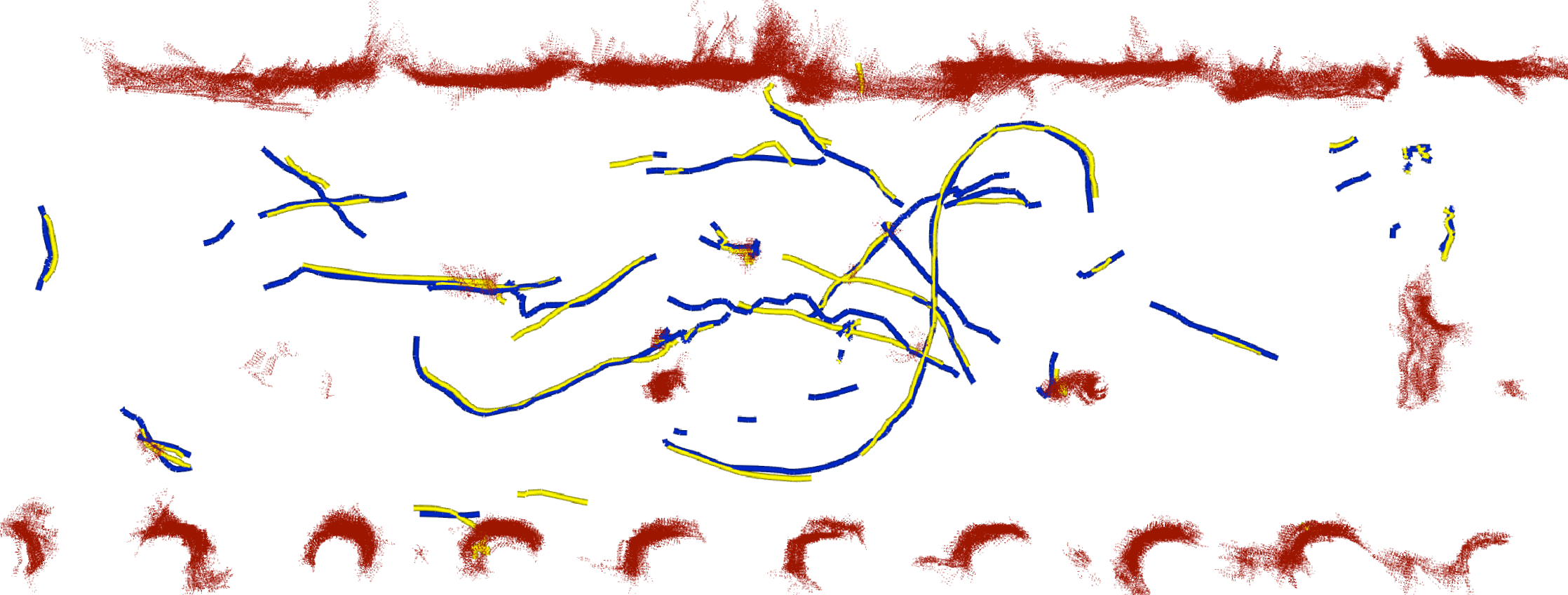}
  \caption{The tracks of dynamic objects (classified by the module presented in Section \ref{classification}) are shown in yellow, the tracks of our ground truth based on LiDAR data in blue, and the stereo camera point cloud of static objects in red.
  The tracks are smooth and hence, suggest the applicability of our solution to motion tracking and prediction.
  Short tracks are caused by the obstacles leaving the limited FOV. The hall is of size $30\times 10m$.}
  \label{fig_tracks}
\end{figure}

\begin{figure}[t]
  \centering
  \includegraphics[width=1.0\columnwidth]{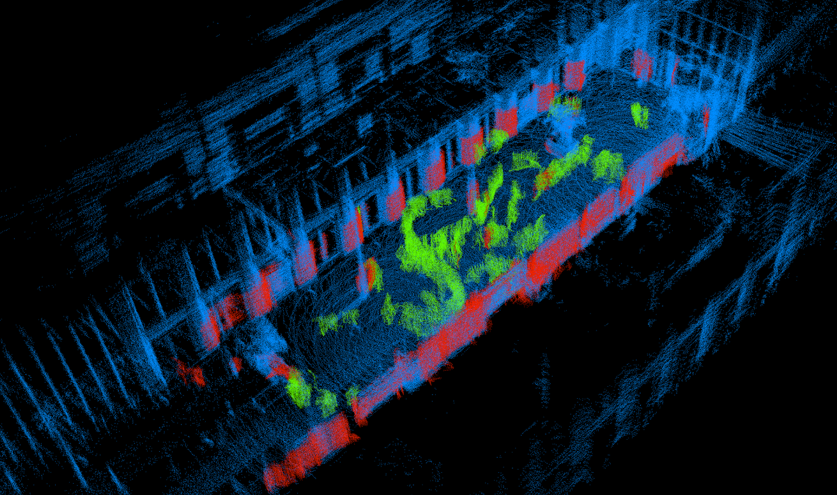}
  \caption{Our static LiDAR ground truth is shown in blue and the accumulated stereo camera point cloud of static objects in red. For better visualization, we also show the accumulated dynamic camera point cloud in green and the LiDAR ground plane, which we remove for the comparison between clouds.}
  \label{fig_LiDAR}
  \vspace{-5mm}
\end{figure}

To evaluate the classification and tracking accuracy of dynamic objects of the proposed system we compare the resulting object tracks to manually labeled tracks from LiDAR measurements that serve as our ground truth. We use the metrics MOTP and MOTA, as defined in the work of Bernardin \cite{bernardin2006multiple}.

For a robotic system navigating in real environments, a precise mapping of static objects matters. As MOTP and MOTA do not evaluate the detection of static objects we evaluate the similarity between the stereo-based and ground truth static clouds.

\subsubsection{Classification and tracking of dynamic objects}

We first generated a map $m_{l}$ of a controlled, completely static environment $D_1$, using the measurements of the 3D LiDAR and an ICP-based algorithm similar to the work of Dub\'{e}~\cite{dube2017online}. Subsequently, we localized our robot within $m_{l}$ and recorded a second dataset $D_2$, this time with dynamic objects present.
We created a ground truth of obstacle tracks by first manually excluding from $D_2$ all LiDAR measurements that belong to static objects. We then extracted the upper bodies of the pedestrians in the remaining LiDAR clouds by cropping by height, in order to remove the influence of moving legs and hence, to attain smooth trajectories.
Subsequently, we applied Euclidean clustering and tracked the clusters $C^{lidar}$ using closest centroids, in the same manner as presented in Section \ref{clustering_tracking}. We visually inspected and adjusted the resulting LiDAR ground truth tracks ${T_{t, k}^{i, lidar}=\{c_{t-k\cdot \Delta t}^{lidar}, ..., c_{t-\Delta t}^{lidar}, c_{t}^{i, lidar}\}}$ to ensure the absence of false positives, false negatives, or mismatches. $D_2$ is of $4\si{\min}$ length and features $31$ encounters with pedestrians leading to $1267$ ground truth pedestrian positions $c^{lidar}$.
Figure \ref{fig_tracks} shows a top-down view of our LiDAR-based ground truth tracks $T^{i, lidar}$ and the camera-based dynamic object tracks $T^{j}$ extracted from $D_2$ by our proposed system.
To compute the MOTP and MOTA, we set the threshold $l_T$ for a correct match between object (object from the ground truth LiDAR track) and hypothesis (object from the camera track) as $l_T = 0.4m$, assuming a diameter of $0.4m$ for an average person and hence, an incorrect match if the object and the hypothesis do not have any overlap. Using this, we reach a MOTP of $0.07 \pm 0.07m$ and a MOTA of $85.3\%$, which is composed of a false negatives rate $f_n = 8.3\%$ (covering non-detected dynamic objects and dynamic objects erroneously classified as static or uncertain), a false positives rate $f_p = 3.0\%$ (covering ghost objects or static objects misclassified as dynamic), and a mismatch rate $f_m = 3.3\%$.

\subsubsection{Classification of static objects}
\begin{figure}[t]
  \centering
  \includegraphics[width=1.0\columnwidth]{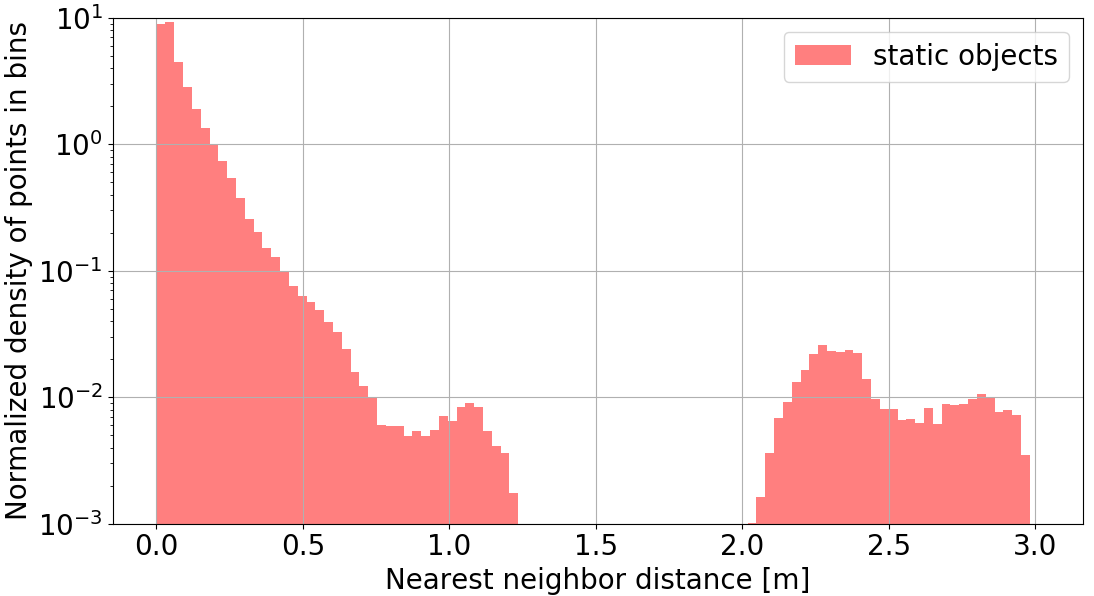}
  \caption{Precision of static classification: normalized histogram of nearest neighbor distances $d$ from the static part of the camera point cloud to our static LiDAR ground truth point cloud.
  }
  \label{fig_classification_overview}
\end{figure}

The map $m_l$ from $D_1$ served as our ground truth for static objects.
We then accumulated all stereo camera clusters classified as static in $D_2$
resulting in the point cloud $m_s$.
Figure~\ref{fig_LiDAR} visualizes $m_l$ overlaid with $m_s$.
To evaluate the similarity between the stereo-based and ground truth static clouds we measure the nearest neighbor distances $d$ from points of the camera cloud $m_s$ to the LiDAR cloud $m_l$.
Ideally, the static camera cloud $m_s$ coincides with the LiDAR ground truth $m_l$.
Figure~\ref{fig_classification_overview} shows the normalized distribution of these distances $d$. Ideally, $d$ is zero for static objects.
However, as shown in Section \ref{exactness_completeness}, in the extreme case static points can differ up to $l_c = 0.8m$ from the static ground truth.
As there is no ground truth available for the per-point-classification, we estimate correct static classifications of our pipeline using one threshold $l_e$. We chose this threshold $l_e$ to be the average of both limit cases of $l_c = 0.8m$ and zero, hence $l_e = 0.4$. We declare all static points below $l_e$ of the camera cloud as correctly classified,
reaching a precision of the classification of static objects of $96.9\%$.

\subsection{Run-time evaluation} \label{timing}

Table \ref{table_timing} shows the timing of the modules of our pipeline when deployed on the embedded system described in Section \ref{experimental_setup}.
The two most significant contributors
are the block-matching and the people-detector network. Overall, the pipeline runs at $8.5\si{\Hz}$ with all features. Excluding the people detector, we reach $13.5\si{\Hz}$. The core part of our work, i.e. point cloud filtering, clustering and 3D tracking, classification, and motion estimation requires $21ms$ per stereo camera frame.

\begin{table}[h]
\caption{Average run-times of the major modules of our dynamic obstacle detection and tracking pipeline on a standard CPU.}
\label{table_timing}
\begin{center}
\vspace*{-4mm}
\begin{tabular}{|c|c|c|}
\hline
\textbf{Module} & \textbf{Timing [ms]} & \textbf{Portion [\%]}\\
\hline
\hline
Point Cloud Generation & 55.2 & 46.4 \\
\hline
2D People Detector & 42.4 & 35.6 \\
\hline
Point Cloud Filtering & 9.9 & 8.3 \\
\hline
Classification as Dynamic or Static & 8.2 & 6.9 \\
\hline
Remaining modules & 3.3 & 2.8 \\
\hline
\end{tabular}
\end{center}
\end{table}

\section{CONCLUSION} \label{conclusion}

In this paper we presented a method that reliably detects and tracks both generic dynamic objects and humans based on stereo images and thus provides accurate perception capabilities enabling compliant navigation in crowded places. Our novel algorithm detects generic dynamic objects based on motion and geometry in parallel to a detector network, which classifies humans as such based on their visual appearance. We handle short-time occlusions by estimating the velocities of the tracked objects and provide a 2D occupancy grid
that is suitable for performing obstacle avoidance. We showed that our system is real-time capable on a computationally-constrained platform and
despite the high noise level of stereo camera data we achieve a MOTP of $0.07 \pm {0.07m}$, and a MOTA of $85.3\%$ for the detection and tracking of dynamic objects, and a precision of $96.9\%$ for the detection of static objects.

Future work will focus on advanced motion models, ground plane analysis and expanding the pipeline to use multiple perception sensors simultaneously.

 \addtolength{\textheight}{-0cm}   %
\bibliographystyle{plain}
\bibliography{bibliography.bib}

\end{document}